\documentclass[journal]{IEEEtran}
\pdfoutput=1

\usepackage{xcolor}

\usepackage{svg}
\usepackage{graphicx}
\graphicspath{{Figures/}}
\usepackage{rotating,siunitx}
\usepackage{hyperref}

\usepackage{pgfplots}
\usepackage{amsmath}
\DeclareMathOperator*{\argmin}{arg\,min}

\usepackage{bm}
\usepackage[normalem]{ulem}
\pgfplotsset{width=10cm,compat=1.9}

\usepackage{caption} 
\usepgfplotslibrary{external}
\begin{document}
\title{An Open Source Design Optimization Toolbox Evaluated on a Soft Finger}

\author{Stefan Escaida Navarro$^{*\dagger,\ddagger}$,
        Tanguy Navez$^{*\ddagger}$,
        Olivier Goury$^{\ddagger}$,
        Luis Molina$^{\dagger}$ and
        Christian Duriez$^{\ddagger}$% <-this % stops a space
\thanks{$^{*}$Authors have contributed equally to this work.$^{\dagger}$Authors are with the Institute for Engineering Sciences of the O'Higgins University, $^{\ddagger}$ authors are with the team DEFROST at Inria Lille - Nord Europe as well as CRIStAL at the University of Lille. This work has been partially funded by the Agencia Nacional de Investigación y Desarrollo (ANID) in Chile with the ``Fondecyt the Iniciación'' grant number 11230505. Tanguy Navez would like to acknowledge the support of both the Region Hauts de France and the ANR through the program AI\_PhD@Lille co-funded by the University of Lille and INRIA Lille - Nord Europe.}% <-this % stops a space
}


\markboth{
 T\MakeLowercase{his work has been submitted to the} IEEE \MakeLowercase{for possible publication.} C\MakeLowercase{opyright may be transferred without notice, after which this version may no longer be accessible.}
}{Header}



\maketitle

\begin{abstract}
In this paper, we introduce a novel open source toolbox for design optimization in Soft Robotics. We consider that design optimization is an important trend in Soft Robotics that is changing the way in which designs will be shared and adopted. We evaluate this toolbox on the example of a cable-driven, sensorized soft finger. For devices like these, that feature both actuation and sensing, the need for multi-objective optimization capabilities naturally arises, because at the very least, a trade-off between these two aspects has to be found. Thus, multi-objective optimization capability is one of the central features of the proposed toolbox. We evaluate the optimization of the soft finger and show that extreme points of the optimization trade-off between sensing and actuation are indeed far apart on actually fabricated devices for the established metrics. Furthermore, we provide an in depth analysis of the sim-to-real behavior of the example, taking into account factors such as the mesh density in the simulation, mechanical parameters and fabrication tolerances.  
\end{abstract}


% \fi
\IEEEpeerreviewmaketitle

\section{Introduction}
\label{sec:Introduction}
\IEEEPARstart{W}{ithin} the domain of Soft Robotics, plenty of novel designs for actuators, mechanisms, and sensors using a variety of materials have been proposed over the last years. The dissemination of these designs happens through a variety of channels: through scientific literature, through lab-to-lab sharing, and online platforms. Examples for such platforms are the Soft Robotics Toolkit\footnote{\url{https://softroboticstoolkit.com/}}, Instructables\footnote{\url{https://www.instructables.com/Soft-Robotics/}}, and our own website for the Soft Robots plugin for SOFA featuring both designs and simulation files\footnote{\url{https://project.inria.fr/softrobot/}}. Some of the designs and fabrication methods have since been adopted widely, such as the~\emph{PneuNet} actuator~\cite{shepherd_multigait_2011} or the \emph{Vine Robot}~\cite{hawkes2017soft}. 

\begin{figure}
    \centering
    \small
    \includegraphics[width=\linewidth]{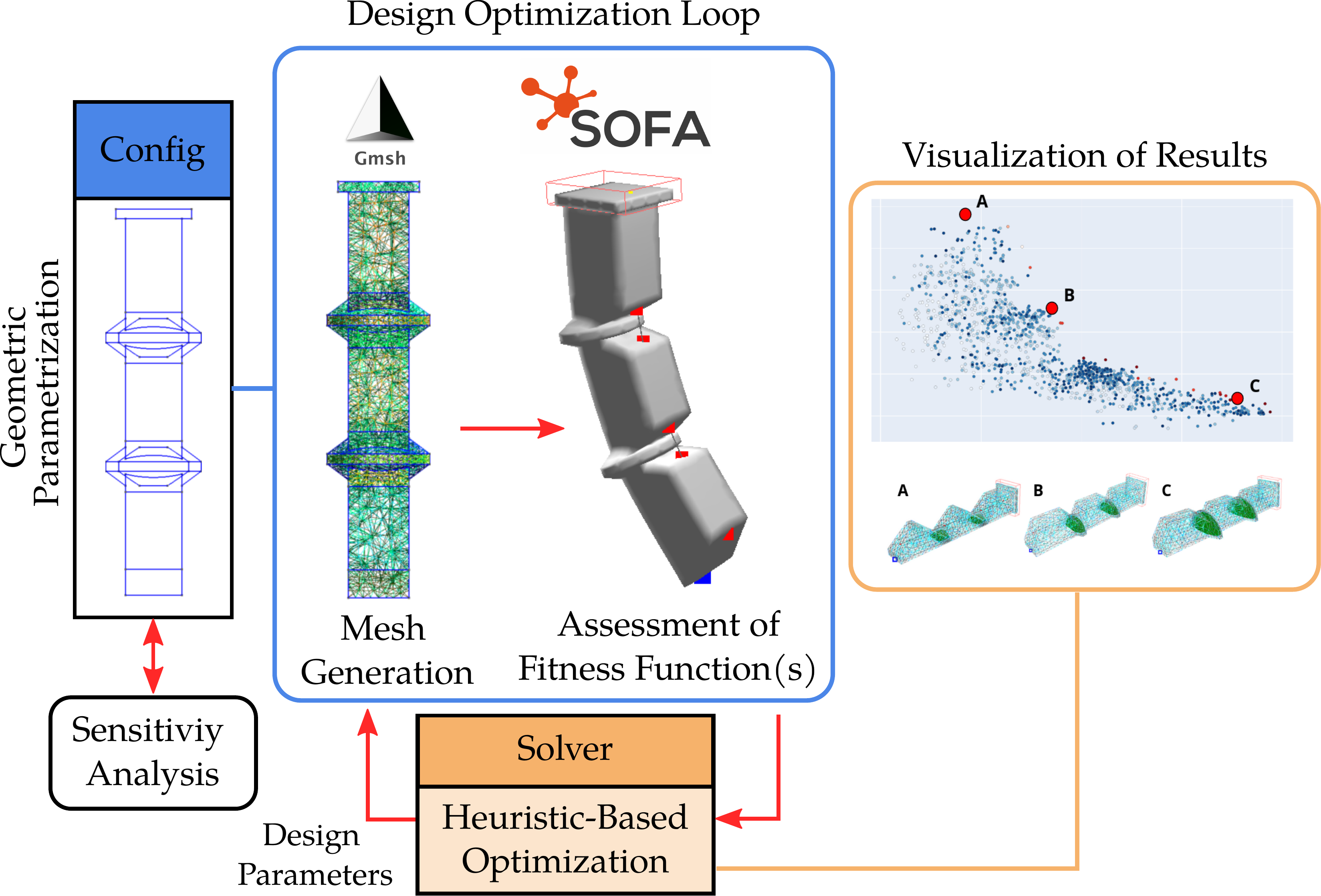} % , natwidth=1169,natheight=672
     \caption{The Design Optimization Toolbox is based on fully parametric description of soft robotics devices coupled to the simulation framework SOFA. Users can implement their own \emph{Config} class describing a generic design and multiple fitness functions. Volumetric meshes are generated using the Python API of \emph{Gmsh}, allowing to evaluate the optimization objectives within a SOFA simulation. A \emph{sensitivity analysis} feature enables exploring relationships between objectives and design parameters. Heuristic-based algorithm are implemented through the \emph{Solver} class for efficiently exploring the Pareto Front in the design space. }
\label{fig:Intro}
\end{figure}

The available resources allow users to replicate the designs, oftentimes with step-by-step tutorials. Despite this, if they want to adapt a design for their own specific application, they need to get back to the metaphorical drawing board to redesign the parts and everything that is needed for fabrication (molds, laser-cut parts, etc.). However, considering the trend towards computational design in Soft Robotics, this is no longer an up-to-date mindset. The mindset that is starting to emerge is to think of families of designs that are given by a parameterized representation of the devices of interest. Integrated into this way of thinking are simulation tools that are fed by the parametric designs allowing to find optimal ones computationally. This is the type of architecture that we propose in Fig.~\ref{fig:Intro}. 

The goal of this paper is to make contributions that trace the path towards the mindset referred to above on the example of a sensorized soft finger. These contributions come in the form of an open source design optimization toolbox and accompanying tutorials. The first step in the paper is to present a completely parametric version of the soft finger that can readily be simulated with SOFA. For describing the parametric designs and generating the meshes needed for simulation, we use the Python API for the free software \emph{Gmsh}~\cite{geuzaine2009gmsh} (see also Fig.~\ref{fig:Intro}). The 3D-printed molds necessary for the silicone casting steps during the fabrication are derived automatically for the design parameters. The second step is to present a \emph{design optimization toolbox}, capable of multi-objective optimization. This capability is showcased in a scenario where we simultaneously optimize two central aspects of a soft finger's design: its workspace, characterized by the \emph{angular displacement}, and the sensitivity to deformation of the embedded fluidics sensors, characterized by the~\emph{deformation volume}. In the third step, we go on to fabricate instances of the soft fingers developed with the presented tools and take an in depth look at the question of how well the optimization results predict performance in reality, i.\,e.\ an analysis of the sim-to-real transfer. The code for the toolbox as well as the fabrication tutorials are available as part of a repository for SOFA called \emph{SoftRobots.DesignOptimization} plugin. \footnote{\url{https://github.com/SofaDefrost/SoftRobots.DesignOptimization}}

The remainder of the paper is structured as follows: In the next section, we discuss the related work from the field. Then, we present the concept for our design optimization toolbox. In Sec.~\ref{sec:OptimalDesignFinger}, we show the optimization results for the proposed finger and in Sec.~\ref{sec:ExperimentalValidation} we further validate the results with experiments using fabricated fingers. Finally, in Sec.~\ref{sec:Conclusions}, we provide a summary of the contributions and give a perspective for future works. 

\section{Related Works}
\label{sec:RelatedWorks}

%Soft-robots are still primarily designed by hand through replacing hard parts by soft ones in rigid robots or reproducing what can be found in nature. Thus, 
There is a growing interest from the Soft Robotics community for tools enabling efficient design exploration. However, most of the available tools are still restricted to a few soft robot modeling strategies or to specific design applications. In \cite{yao2022simulation}, a toolbox for exploring design of a standardized soft pneumatic actuators based on FEA simulation is introduced. However, this toolbox is limited to a single parametric design and lacks the integration of optimization tools for efficiently exploring the design space. A more complete toolbox for simulating, optimizing the design and controlling soft robots has been proposed in~\cite{mathew2022sorosim}. However, it is based on the Geometric Variable-Strain model of Cosserat rod which constrains its application to robots with rod-like shapes. There is a need for a generalized modeling platforms enabling to tackle applications in design optimization and model-based control without the need to develop several specific scripts in different software frameworks.

%in the domain of design optimization %
% On the one hand, 
One challenge is the choice of mathematical representation of the design variables. Some works are interested in design parameters encoding in discretized spaces, for instance on a voxel grid. Design variables such as materials are then optimized at the scale of each voxel. Although these kind of approaches enable to generate almost free geometries, they still lack efficient soft actuators modelling and struggle to integrate manufacturability constraints. A comprehensive review of works in both topology optimization and generative approaches applied to soft robot design optimization is proposed in~\cite{pinskier2022bioinspiration}. 
Other works use high-level design parameters encodings consider the robot as an association of modular components~\cite{schaff2022soft} or geometrical components of variable sizes~\cite{yao2022simulation}~\cite{morzadec2019toward}. These types of encoding rely on the designer's expertise in designing an initial geometric modelling of the robot. Then, optimization is performed for variation around this initial model. The result is a more restricted design space, but one that can more easily ensure compliance with physical fabrication constraints as well as an acceptable optimization time. %In this work, we consider the optimization of generic design with a focus on both manufacturability and simulation-to-reality transfer.

In~\cite{schaff2022soft}, the authors propose a \emph{co-optimization} of the control policy and the design of a crawling robot, using SOFA. They discuss that the optimization outcome outperforms an expert designed device. The robot is a disc with eight ports where pneumatic legs can be attached. The design space is thus discrete. The authors chose to employ model-order-reduction~\cite{goury2018fast} to make the exploration of the control/design space feasible in terms of time effort. This works highlights the feasibility of using SOFA for optimization tasks that successfully transfer to real robot platforms. In~\cite{pingchuan2021diffaqua}, Ma et al.\ introduce their work on optimizing shape and control of swimmers using a differentiable simulation, which is leveraged for a gradient-based optimization approach. Their work considers both co-optimization of shape and control as well as multi-objective optimization of swimming speed and energy efficiency. An optimal design is found by interpolating between previously established shapes. Thus, their design space is continuous, but limited to barycentric combinations of the initial designs. Co-design of both soft robot morphology and control is also discussed in~\cite{wang2023so} where several design representations and optimization algorithms are evaluated on a benchmark of locomotion tasks. In~\cite{tapia2020makesense}, Tapia et al.\ present an approach for optimal sensorization of soft robots using stretch sensors to reconstruct their configuration. The robot's mechanics are modeled using the FEM and a model for the resistive type sensors is established, allowing to relate deformations to sensor values and vice-versa. They also simulate the actuation of the robot by means of pneumatic chamber. However, the authors do not report optimizing for both actuation and sensing. Finally, in~\cite{spielberg2021co} Spielberg et al.\  report on a method for co-learning task and sensor placement for soft robots. Their work highlights the importance of optimizing for sensor placement, as this impacts the ability to properly reconstruct the state of the robot from the sparse sensor data. However, their work does not address the challenge of optimizing the shape of the robots and they do not evaluate the simulation to reality transfer.

% In summary, a variety of methods exist in literature to optimize the design of soft robots, oftentimes co-optimizing control policies. The choice of design variables (discrete, continuous, interpolated, etc.) as well as the considerations of fabrication constraints are important. The challenge of sensorization is also addressed in a few cases. However,

% Although we evaluate the toolbox on the example of a soft finger in this article, we aim to manage completely generic designs using the whole feature set of SOFA available for exploring different modes of actuation, sensing, including contacts and functionality such as model-order reduction. 

In summary, to our knowledge, a multi-objective optimization of actuation and sensing has not been presented yet. Also, while some of the approaches include useful features for optimization, such as differentiable simulations, the design space is often discrete or reduced to combinations of existing designs. In addition, we provide more insight for the sim-to-real behavior compared to previous works.
\section{Design Optimization Toolbox}
\label{sec:ComputationalDesignToolbox}
The introduced toolbox is built as an interface between the SOFA Framework that provides the simulation and Python libraries for scriptable design and automatic meshing with Gmsh as well as heuristic-based optimization, as illustrated by Fig.~\ref{fig:Intro}. Design optimization is done through the exploration of design parameters with regards to user-defined objective functions. The mechanical behavior of soft robot deformations can be described through continuum mechanics for which there are no analytical solutions in the general case. The toolbox enables the use of many simulation functionalities developed around the SOFA Framework to simulate and evaluate a given design, including actuation~\cite{coevoet2017software} and sensor~\cite{navarro2020model} models, different modeling strategies (Finite Element Method (FEM), Cosserat Rod~\cite{adagolodjo2021coupling}) for computing numerical solutions, inverse solvers and contact modeling~\cite{coevoet2017optimization}. SOFA has in fact already been used for design optimization tasks~\cite{schaff2022soft,morzadec2019toward}, as discussed in Sec.~\ref{sec:RelatedWorks}. In order to be able to use the FEM for modeling a soft robot, we introduce tools for automatically regenerating meshes when the design parameters evolve in sec.~\ref{subsec:ParametricDesignByScripting}.
%for design optimization
% In this section, we describe the main classes and features of the project as described in.
% Thanks to the modularity of the proposed toolbox, new scripts taking advantage of Python computation libraries are easily implemented. 
The proposed toolbox is developed in an object-oriented programming fashion. The user can easily define his own components inheriting from  base classes in a well-structured manner. For each class, one to several component extensions are provided as an example. Following the scheme of Fig.~\ref{fig:Intro}, a design optimization problem is represented through the \emph{Config} class. It describes the design parameters and their bounds. This class makes the link with user-defined scripts both for generating new design meshes and for evaluating their fitness function(s) in \emph{SOFA} simulation scene(s). The \emph{Solver} class manages the design optimization and results visualization. %Note that although we focus on exploring soft robot designs in this paper, this toolbox could also be used for non-robotics related applications, such as calibrating design and simulation parameters of rigid or deformable structures.

\subsection{Parametric Designs by Scripting}
\label{subsec:ParametricDesignByScripting}
\begin{figure}
    \centering
    \includegraphics[width=0.85\linewidth]{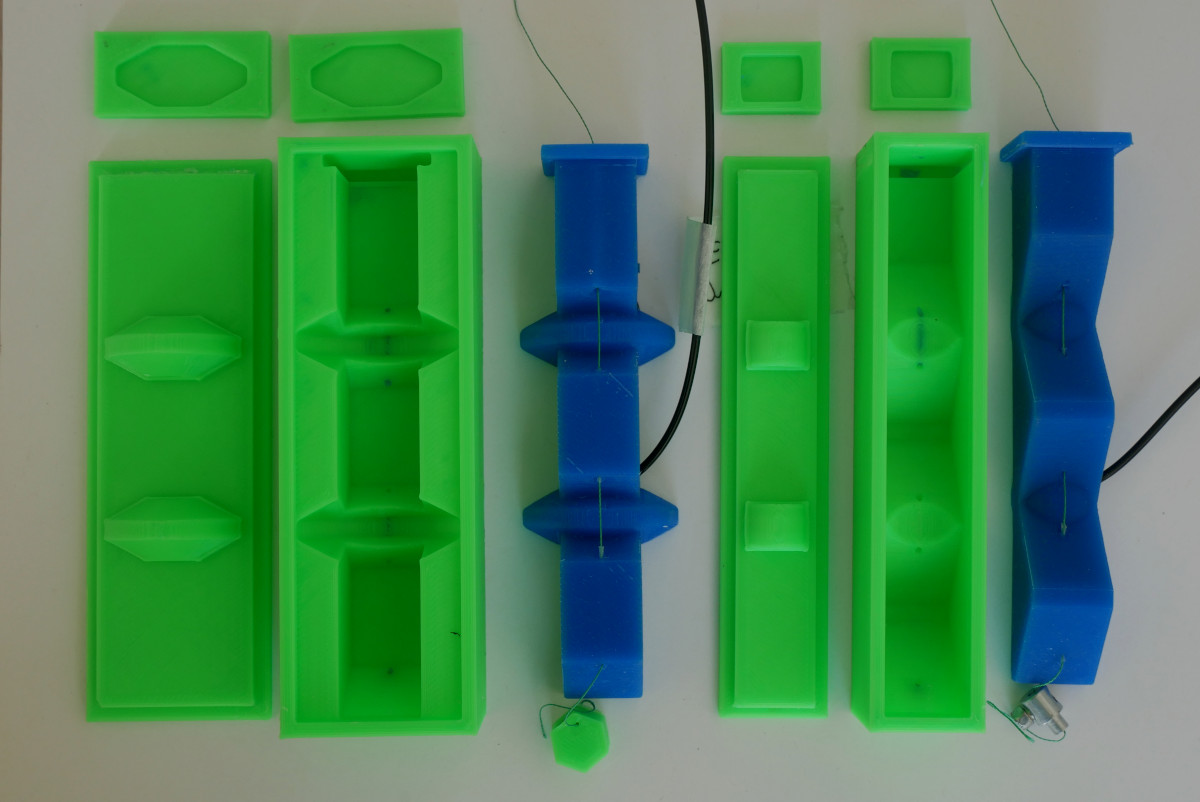}
    \caption{Both fingers, large and slim, fabricated using the corresponding automatically generated molds.}
    \label{fig:Molds+Fingers}
\end{figure}

 The free software Gmsh implements a Python API for the one of its backends, the \emph{Open Cascade Technology} (OCCT). This means the user can create arbitrary 3D shapes by building them in a bottom-up fashion starting with elementary entities, i.\,e.\ from points to curves to surfaces and finally volumes. In addition, 2D and 3D meshes can be generated from the resulting 3D shape with Gmsh. They are thus readily available for simulation within SOFA or other platforms. %As part of the toolbox, we propose a tutorial on creating a pawn with an accordion-shaped body that covers all these steps. 
 We furthermore consider the fabrication process as a part of the computational design phase. The fabrication of the devices based on casting involves several steps. The corresponding molds can therefore be generated automatically from the given finger design to ease the fabrication process, as is shown in the example of the Soft Finger (see Fig.~\ref{fig:Molds+Fingers}). This automation aid to the fabrication process also helps in the dissemination of the designs.
  
\subsection{Design Optimization Framework}
In this section, we describe the main methods of the toolbox, which are summarized by Fig.~\ref{fig:Intro}. The toolbox enables tackling optimization problem of the form:
\begin{equation}
\begin{aligned}
\argmin_{\bm{x}} \quad &(f_1(\bm{x}), f_2(\bm{x}), ..., f_n(\bm{x})) \\
s.t. \quad &a_i < x_i < b_i \quad \forall x_i \in \bm{x} 
\end{aligned}
\end{equation}
where the integer $n$ is the number of objectives and $x_i$ are continuous design variables bound by box constraints. In the case of multi-objective optimization, it is usually not possible to find a solution minimizing all the objective functions. In that case, we are interested in Pareto optimal solutions i.\,e. solutions that cannot be improved regarding any objective function $f_i$ without deteriorating their result with regards to at least one of the other objective functions.

When faced with a design problem with many variables, design sensitivity analysis could be performed as a pre-processing step. It computes the gradient of the fitness functions with respect to the design variables. They are normalized and visualized as a graph so that the user can easily select the most relevant design variables during optimization as shown with the example of the Finger (see Fig~\ref{fig:SensitivityAnalysis}).

Once the problem is properly refined, we can enter the optimization loop. It makes the link between the solver, the design meshes generation and the analysis of the fitness function(s) in simulation. Our implementation is fully parallelized in order to improve the computation time. The simulations are run in a \emph{headless} mode, i.\,e.\ without GUI. In the current version, we provided a linking with the open source library \emph{Optuna}~\cite{akiba2019optuna} for the \emph{Solver} class. \emph{Optuna} implements multiple standard heuristic-based optimization algorithms both for single and multi-objectives optimization. It also offers database management as well as results visualization tools. Finally, the results can be visualized as a graph. In the case of multi-objective optimization, the user can choose the design among those located on the Pareto Front with the best compromise regarding the different fitness functions.

\section{Optimal Design of a Sensorized Finger}
\label{sec:OptimalDesignFinger}

In this section, we describe how the proposed toolbox is used to optimize a soft finger that we wish to use to build a soft gripper in the future. The current design is based on a previously work~\cite{navarro2020model}. We consider this task to be a proof of concept for the use of the toolbox that showcases the optimization and trade-off of performance in terms of workspace and sensing of such a finger. For the mentioned task of building a gripper, additional design objectives could be of interest. Considering the action of displacing the cable by a unit amount of $\SI{10}{\milli\meter}$, we want to find the design achieving the largest variation of the finger's cavities volume, for sensing purposes, and at the same time the largest bending angle (for dexterity purposes). These objectives are called \emph{angular displacement} and \emph{deformation volume}, respectively. This could be considered as a trade-off between sensor efficiency and finger dexterity. Definition of the angular displacement is provided in Fig.~\ref{fig:FitnessFunctionsSensorizedFinger}.

\begin{figure}
\centering
\def\svgwidth{0.8\linewidth}
\small
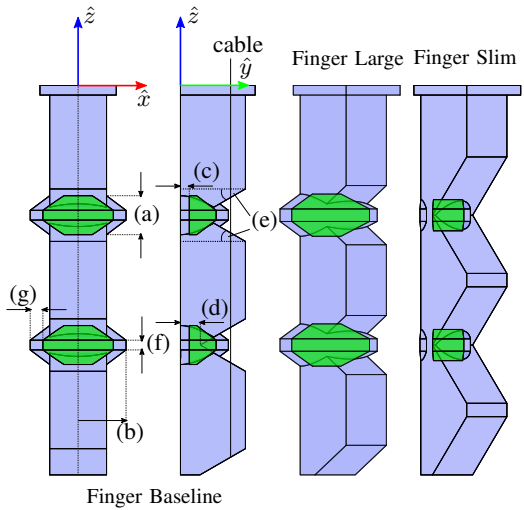
 \caption{Left: The parameters subject to optimization are the following: (a) Cavity Height, (b) Outer Radius, (c) Cork Thickness, (d) Joint Height, (e) Joint Slope Angle, (f) Plateau Height, (g) Wall Thickness. Right: The two instances of the finger selected for evaluation, the \emph{finger large} and \emph{finger slim}.}
\label{fig:FingerDrawing}
\end{figure}

The design parameters subject to change during optimization of the finger are shown in~Fig.~\ref{fig:FingerDrawing} on the left, on the example of our initial design \emph{finger baseline}. All other parameters remain constant, in particular, the thickness, height and length of the finger. The cable is at a fixed distance from the outer radius of the cavity. Thus, the cable passes closer to the center of the finger when a small outer radius is chosen for the cavities.

% In the following, we describe the mathematical formulation of the design objectives and results of the optimization in terms of a Paretto front.

\subsection{Fitness Functions (Design Objectives)}

\begin{figure}
    \centering
    \includegraphics[width=0.95\linewidth]{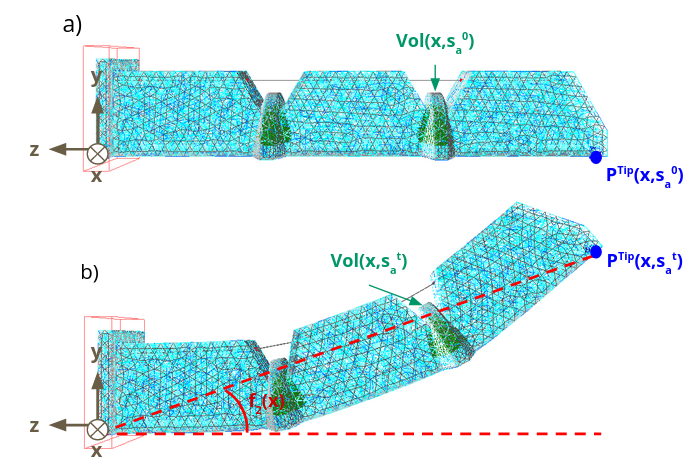}
    \caption{Illustration of the optimized fitness functions. a) Initial equilibrium of the soft finger. b) Equilibrium of the soft finger for an imposed actuation displacement $s_a^{10}$.}
    \label{fig:FitnessFunctionsSensorizedFinger}
\end{figure}

For simulation, we use the FEM model together with the constraint-based cable and pneumatic cavities models from the Soft Robots plugin~\cite{coevoet2017software}. The assessment of the fitness functions is made for the model at static equilibrium. The fitness functions for the quantifying the deformation volume, $f_1$, and the angular displacement, $f_2$ are, respectively, given by:
\begin{equation}
\begin{aligned}
f_1(\bm{x}) &= |Vol(\bm{x}, s_a^{10}) - Vol(\bm{x}, s_a^0)|\\
f_2(\bm{x}) &= \frac{|\arccos(|P^{Tip}_z(\bm{x}, s_a^{10})|)}{||P^{Tip}(\bm{x}, s_a^{10})||}
\end{aligned}
\label{eq:fitnessFunctions}
\end{equation}
where $\bm{x}$ is the set of design variables describing the assessed design. Here, $\bm{x}$ is of dimension $7$ and is displayed in Fig.~\ref{fig:FingerDrawing}, $s_a^0$ and $s_a^{10}$ are the actuation displacements imposed on the cable actuators of respectively \SI{0}{\milli\meter} and \SI{10}{\milli\meter}. The operator \emph{Vol()} gives the volume of a cavity and $P^{Tip}$ is the position of the tip of the finger. The function $f_1$ then computes the deformation volume (the variation of volume) in the cavity for a controlled actuator displacement, which corresponds to what a flow sensor can measure (see Sec.~\ref{sec:ExperimentalValidation}), whereas the function $f_2$ is the angular displacement reached by the fingertip under the same actuator displacement constraints. Notations are introduced in Fig.~\ref{fig:FitnessFunctionsSensorizedFinger}.

\subsection{Optimization Results}
\label{subsec:OptimizationResults}

Simulating the sensorized finger is performed using an Euler implicit time integration method with a time step of 0.01 seconds. Other detailed parameters are available in the open source code on GitHub (see Sec.~\ref{sec:Introduction}).
For the optimization, we made assumptions about the mechanical properties of the finger. We used a Young's Modulus of \SI{3}{\mega\pascal} and a Poisson's Ratio of 0.30, as well as damping coefficient parameters related to mass and stiffness of 0.1. For a given set of design parameters, the objectives are assessed using a 3D mesh having about 500 nodes. With this settings, the simulation takes around 2 seconds to reach equilibrium for each design on our setup\footnote{Laptop with height cores 2.70 GHz Intel Core i7 - 6820}. 

\begin{figure}
    \centering
    \includegraphics[width=0.95\linewidth]{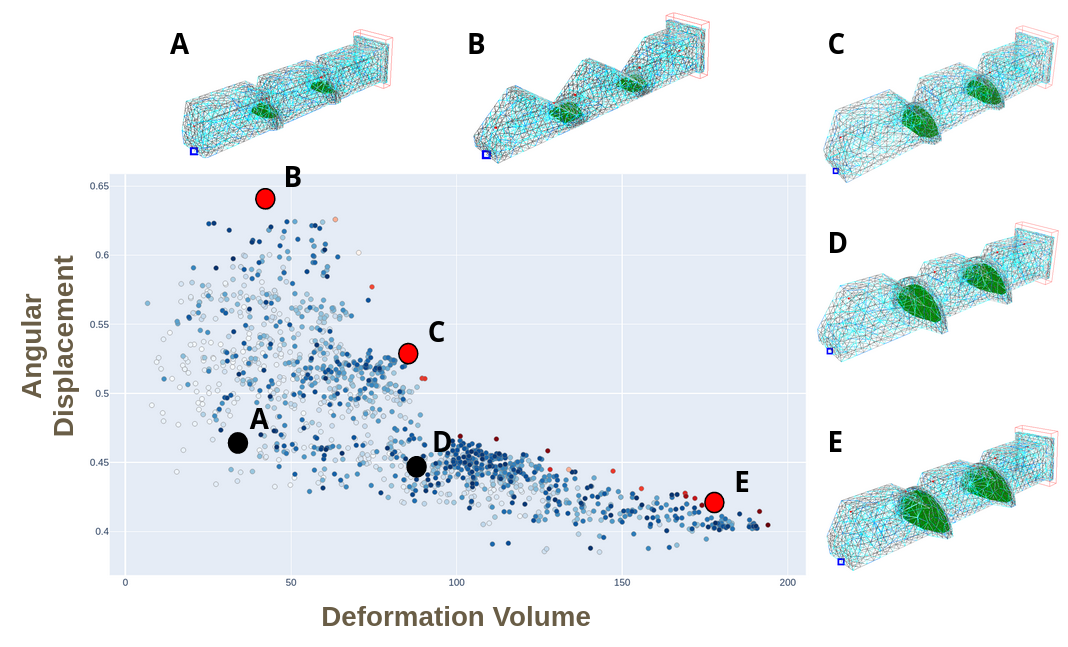}
    \caption{Pareto Front obtained as well as geometries of a few sampled design for the Sensorized Finger design optimization. Results are generated using NSGA-II algorithms with initial population of 50 design candidates, probability of crossover of 0.9, probability of swapping parameters between parents of 0.5.} Each point is the evaluation of a design in simulation. A total of 1500 designs were considered. The Pareto optimal solutions are represented by red dots.
    \label{fig:ResultsFingerPareto}
\end{figure}

Later, in Sec.~\ref{subsubsec:ParetoFrontVariability}, it is discussed more in detail that the results obtained seem to generalize well to other mechanical parameters (Young's Modulus and Poisson's Ratio) and discretization sizes. We use the implementation of the NSGA-II multi-objective evolutionary algorithm from the \emph{Optuna} library as a solver. Algorithm hyper-parameters and results are displayed in Fig.~\ref{fig:ResultsFingerPareto}.

The most performing results in Fig.~\ref{fig:ResultsFingerPareto} with respect to the angular displacement objective are geometries with less material and cavities flattened along the finger length such as design B. The volume of the cavities is then minimal, thus facilitating a more significant deflection of the finger, because the cable can pass closer to the joints. Conversely, the best results regarding the sensitivity to deformation volume consider the largest possible cavities having thin walls, as in design E. The Pareto front enables for efficiently exploring the design space and finding compromises like that of design C.

\subsection{Considerations on the Choice of the Fitness Functions}
\begin{figure}
    \centering
    \includegraphics[width=0.95\linewidth]{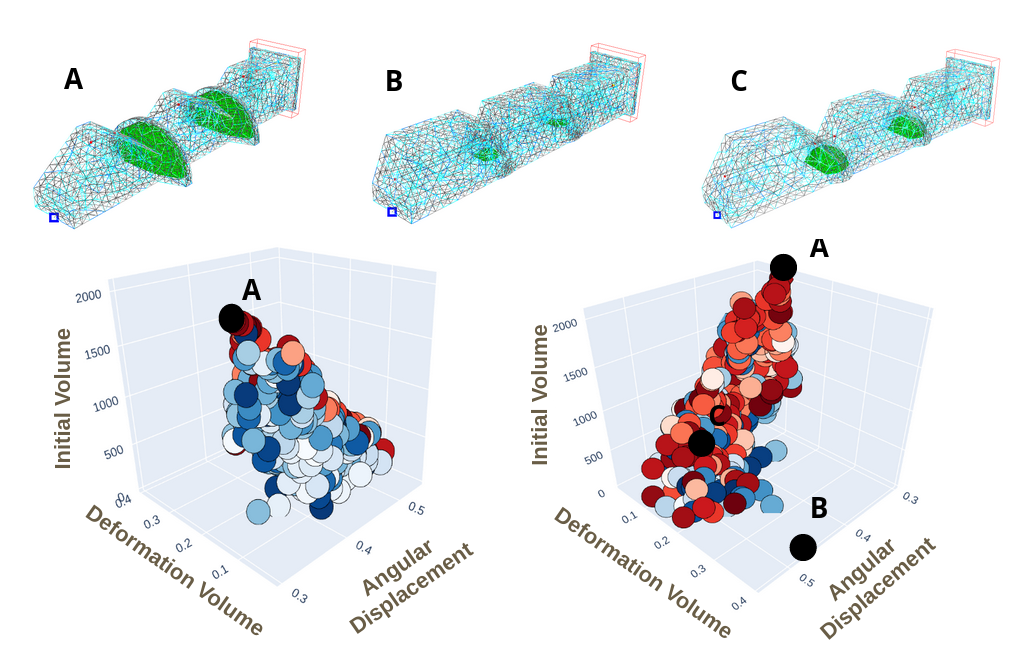}
    \caption{Pareto Front obtained from the design optimization of the Sensorized Finger for using a pressure sensor. Geometries of a few sampled design are also shown: \textbf{A)} $f_2(\bm{x}_A) = \SI{17.2}{\degree}$, $f_3(\bm{x}_A) = 0.085$, $f_4(\bm{x}_A) = \SI{2043}{\micro\liter}$, \textbf{B)} $f_2(\bm{x}_B) = \SI{26.9}{\degree}$, $f_3(\bm{x}_B) = 0.395$, $f_4(\bm{x}_B) = \SI{47}{\micro\liter}$, \textbf{C)} $f_3(\bm{x}_C) = \SI{26.9}{\degree}$, $f_3(\bm{x}_C) = 0.132$, $f_4(\bm{x}_C) = \SI{572}{\micro\liter}$.
}
    \label{fig:ResultsFingerParetoPressure}
\end{figure}

The choice of fitness function has a big impact in the optimization outcome. If pressure measurements are used rather than air-flow sensors (see Sec.~\ref{sec:ExperimentalValidation}), the optimal geometries are quite different. To test this, we designed alternative fitness functions as follows:
\begin{equation}
\begin{aligned}
f_3(\bm{x}) &= \frac{f_1(\bm{x})}{Vol(\bm{x}, s_a^{10})}\\
f_4(\bm{x}) &= Vol(\bm{x}, s_a^0)
\end{aligned}
\end{equation}
where $f_3$ characterizes the change of pressure of the pneumatic chamber under imposed displacement and $f_4$ its initial volume. The resulting Pareto Front is shown in Fig.~\ref{fig:ResultsFingerParetoPressure}.

This time, contrary to the results obtained when evaluating the deformation volume, the flatter the pneumatic cavity, the more effective is the pressure change. The $f_4$ objective on the initial volume of the cavity was added to avoid having pneumatic cavities that are too small and impossible to manufacture. This example demonstrates the relevance of using the provided toolbox for generating designs in accordance with different physical specifications.

\section{Experimental Validation}
\label{sec:ExperimentalValidation}
\begin{figure}
\centering
\def\svgwidth{0.4\linewidth}
\small
%% Creator: Inkscape inkscape 0.92.5, www.inkscape.org
%% Accompanies image file 'ExperimentSetup.pdf' (pdf, eps, ps)
\begingroup%
  \makeatletter%
  \providecommand\color[2][]{%
    \errmessage{(Inkscape) Color is used for the text in Inkscape, but the package 'color.sty' is not loaded}%
    \renewcommand\color[2][]{}%
  }%
  \providecommand\transparent[1]{%
    \errmessage{(Inkscape) Transparency is used (non-zero) for the text in Inkscape, but the package 'transparent.sty' is not loaded}%
    \renewcommand\transparent[1]{}%
  }%
  \providecommand\rotatebox[2]{#2}%
  \newcommand*\fsize{\dimexpr\f@size pt\relax}%
  \newcommand*\lineheight[1]{\fontsize{\fsize}{#1\fsize}\selectfont}%
  \ifx\svgwidth\undefined%
    \setlength{\unitlength}{304.00002547bp}%
    \ifx\svgscale\undefined%
      \relax%
    \else%
      \setlength{\unitlength}{\unitlength * \real{\svgscale}}%
    \fi%
  \else%
    \setlength{\unitlength}{\svgwidth}%
  \fi%
  \global\let\svgwidth\undefined%
  \global\let\svgscale\undefined%
  \makeatother%
  \begin{picture}(1,1.57894719)%
    \lineheight{1}%
    \setlength\tabcolsep{0pt}%
    \put(0,0){\includegraphics[width=\unitlength,page=1]{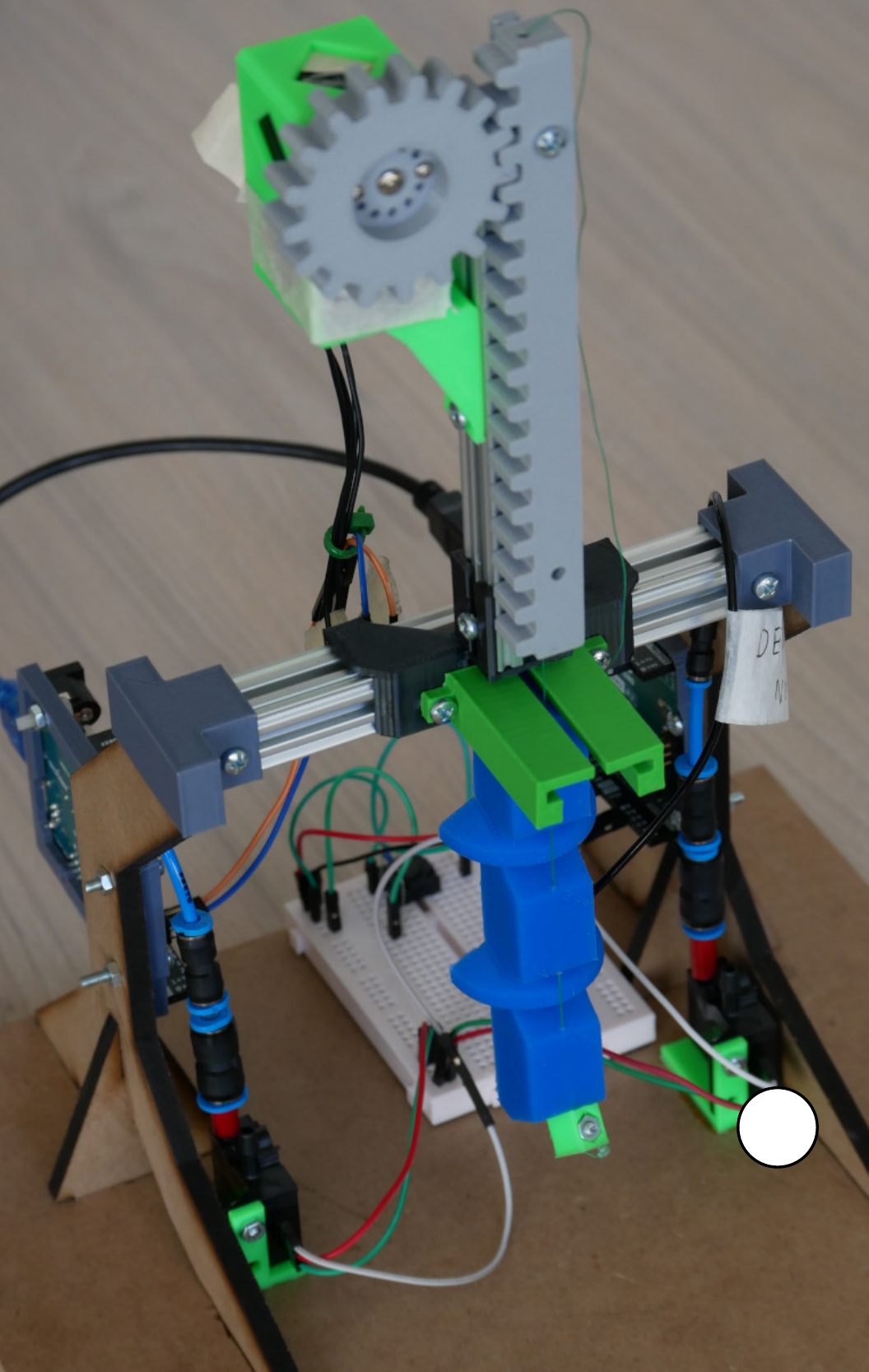}}%
    \put(0.87145968,0.26481586){\color[rgb]{0,0,0}\makebox(0,0)[lt]{\lineheight{1.25}\smash{\begin{tabular}[t]{l}a\end{tabular}}}}%
    \put(0,0){\includegraphics[width=\unitlength,page=2]{ExperimentSetup.pdf}}%
    \put(0.47619179,1.1480283){\color[rgb]{0,0,0}\makebox(0,0)[lt]{\lineheight{1.25}\smash{\begin{tabular}[t]{l}b\end{tabular}}}}%
    \put(0,0){\includegraphics[width=\unitlength,page=3]{ExperimentSetup.pdf}}%
    \put(0.35095191,0.48754978){\color[rgb]{0,0,0}\makebox(0,0)[lt]{\lineheight{1.25}\smash{\begin{tabular}[t]{l}c\end{tabular}}}}%
  \end{picture}%
\endgroup%

 \caption{Test bench for evaluating the finger's performances: (a) Air-flow sensor, (b) linear actuator and (c) reset button for integration.}
\label{fig:ExperimentSetup}
\end{figure}

The test-bench used to validate the performance of the fingers is shown in Fig.~\ref{fig:ExperimentSetup}. For all experiments, we are interested in the effect of a displacement of $\SI{10}{\milli\meter}$ of the cable, as discussed in Sec.~\ref{sec:OptimalDesignFinger}. To evaluate deformation volume, we have attached an air-flow sensor D6F-P0001A1 by the OMRON Corporation\footnote{\url{https://omronfs.omron.com/en_US/ecb/products/pdf/en-D6F_series_users_manual.pdf}} to the distal cavity. The analog signal of the sensors is digitized using an Arduino UNO board. On the same board, the time integration is done to obtain the volume measurements from the flow. This technique allows measuring very small volumes and exhibits high repeatability, as previously reported in~\cite{navarro2019modeling}. We did not look into using pressure sensors here, because the tubing and the sensor itself represent a rigid volume that needs to be taken into account to obtain volume estimates. However, determining this rigid volume is a more involved process whereas the measurements of the air-flow sensors can be interpreted directly as is. To measure the angular displacement, we used photographs, as shown in Fig.~\ref{fig:ComparisonAngularDisplacement}.  

For validating the results obtained by running the optimization, we have fabricated three instances of the finger. We created one \emph{finger slim} and two identical \emph{finger large} (\#1 and \#2). The idea of fabricating the same design twice was to assess the variability introduced by the fabrication process. In particular, sealing the cavities involves a manual gluing process that we assumed could impact the deformation volume. We first present the results of what is predicted by the simulation in Table~\ref{tab:SensitivitySOFA}. It is shown that a large difference of 515\% exists between the predicted deformation volume of the slim and large fingers. The change in angular displacment is not quite as dramatic, but still clearly noticeable, i.\,e.\ an increase of \SI{9.03}{\degree} or 147\% from large to slim.

\begin{table}
\centering
\begin{tabular}{||c|c|c||} 
 \hline
 Device &  Deform.\ Vol.& Ang.\ Disp. (deg.) \\ [0.5ex] 
\hline\hline
Finger Baseline & $\SI{41.5}{\micro\liter}$ & 21.1 \\ 
\hline
Finger Slim & $\SI{29.7}{\micro\liter}$ & 28.13 \\ 
 \hline 
Finger Large & $\SI{153.1}{\micro\liter}$ & 19.1 \\ 
\hline
Diff Slim and Large & 515\% & 9.03 (147\%) \\
\hline 
\end{tabular}
\caption{Deformation Volumes in SOFA}
\label{tab:SensitivitySOFA}
\vspace{-2mm}
\end{table}

\subsubsection{Deformation Volume}

\begin{table}
\centering
\begin{tabular}{||c|c|c|c||} 
 \hline
 Device & Deform.\ Vol.  (\si{\micro\liter}) & Std (\si{\micro\liter}) & Std (\%) \\ [0.5ex] 
 \hline\hline
 Finger Slim T1 & 45.7 & 0.50 & 1.10 \\ 
 \hline
 Finger Slim T2 & 45.5 & 0.81 & 1.78 \\
 \hline 
 Finger large \#1 & 300.7 & 2.24 & 0.75 \\
 \hline
 Finger large \#2 T1 & 304.7 & 1.46 & 0.48 \\
 \hline
 Finger large \#2 T2 & 307.9 & 1.75 & 0.57 \\
\hline 
\end{tabular}
\caption{Sensitiviy of Deformation Volume Measurments}
\label{tab:Sensitivity}
\vspace{-5mm}
\end{table}

For assessing the deformation volume, each experiment was repeated 5 times. The mean value, together with the standard deviation is reported for all experiments in Table~\ref{tab:Sensitivity}. In Table~\ref{tab:Metrics}, metrics are reported that allow to compare the experiments. We first remark that the experiments are quite repeatable, where a series of 5 repetitions shows a deviation of less than 1\%, except for the slim finger, where the value is slightly higher. This is probably because the overall measured volume is small and thus the measurements are more affected by the noise floor. 

 \begin{table}
\centering
\begin{tabular}{||c|c||} 
 % \hline
 % Device & Vol.\ Sensitivity (\si{\micro\liter}) & Std (\si{\micro\liter}) & Std (\%) \\ [0.5ex] 
 
 \hline
 \hline
 Metric & Value \\
 \hline
Gain slim to large (sim.) & 5.15 (515\%) \\
 \hline
Gain slim to large \#2 & 6.74 (674\%) \\
 \hline
 Diff finger large \#1 vs \#2 &  \SI{7.21}{\micro\liter} (2.40 
 \%) \\
 \hline
 Diff slim T1 vs T2 &  \SI{0.22}{\micro\liter} (0.49 \%) \\
 
 \hline
 Diff large T1 vs T2 &  \SI{3.25}{\micro\liter} (1.06 \%) \\
 \hline
 \hline 
\end{tabular}
\caption{Metrics for comparisons}
\label{tab:Metrics}
\vspace{-5mm}
\end{table}

When comparing the slim finger with the large finger, we observe in Table~\ref{tab:Metrics} that the gain in Deformation Volume is 6.74 (674\%), which is a quite significant increase and is also a figure similar to what was predicted by SOFA, which is a gain of 5.15 (515\%). When we compare the two identical large fingers, we observe a difference of \SI{7.21}{\micro\liter} or $2.4\%$. This indicates that the variability in deformation volume introduced by fabrication is much lower than the variability due to design. Finally, when comparing the results of the exact same fingers (slim and large \#2) after remounting the fingers into the test-bench (trial 1 and trial 2), we observed a variability of \SI{0.22}{\micro\liter} (0.49\%) and \SI{3.25}{\micro\liter} (1.06\%), respectively. We therefore conclude that variability due to fabrication and experimental conditions is small compared to the variability due to design. Thus, the optimization process run in simulation has in fact predicted a significant change in performance for the deformation volume on real world devices.

\subsubsection{Angular Displacement Performance}
\begin{figure}
\centering
\def\svgwidth{0.8\linewidth}
\small
%% Creator: Inkscape 1.1.2 (0a00cf5339, 2022-02-04), www.inkscape.org
%% Accompanies image file 'FingerDesignComparison.pdf' (pdf, eps, ps)
\begingroup%
  \makeatletter%
  \providecommand\color[2][]{%
    \errmessage{(Inkscape) Color is used for the text in Inkscape, but the package 'color.sty' is not loaded}%
    \renewcommand\color[2][]{}%
  }%
  \providecommand\transparent[1]{%
    \errmessage{(Inkscape) Transparency is used (non-zero) for the text in Inkscape, but the package 'transparent.sty' is not loaded}%
    \renewcommand\transparent[1]{}%
  }%
  \providecommand\rotatebox[2]{#2}%
  \newcommand*\fsize{\dimexpr\f@size pt\relax}%
  \newcommand*\lineheight[1]{\fontsize{\fsize}{#1\fsize}\selectfont}%
  \ifx\svgwidth\undefined%
    \setlength{\unitlength}{1363.45319132bp}%
    \ifx\svgscale\undefined%
      \relax%
    \else%
      \setlength{\unitlength}{\unitlength * \real{\svgscale}}%
    \fi%
  \else%
    \setlength{\unitlength}{\svgwidth}%
  \fi%
  \global\let\svgwidth\undefined%
  \global\let\svgscale\undefined%
  \makeatother%
  \begin{picture}(1,1.20442311)%
    \lineheight{1}%
    \setlength\tabcolsep{0pt}%
    \put(0,0){\includegraphics[width=\unitlength,page=1]{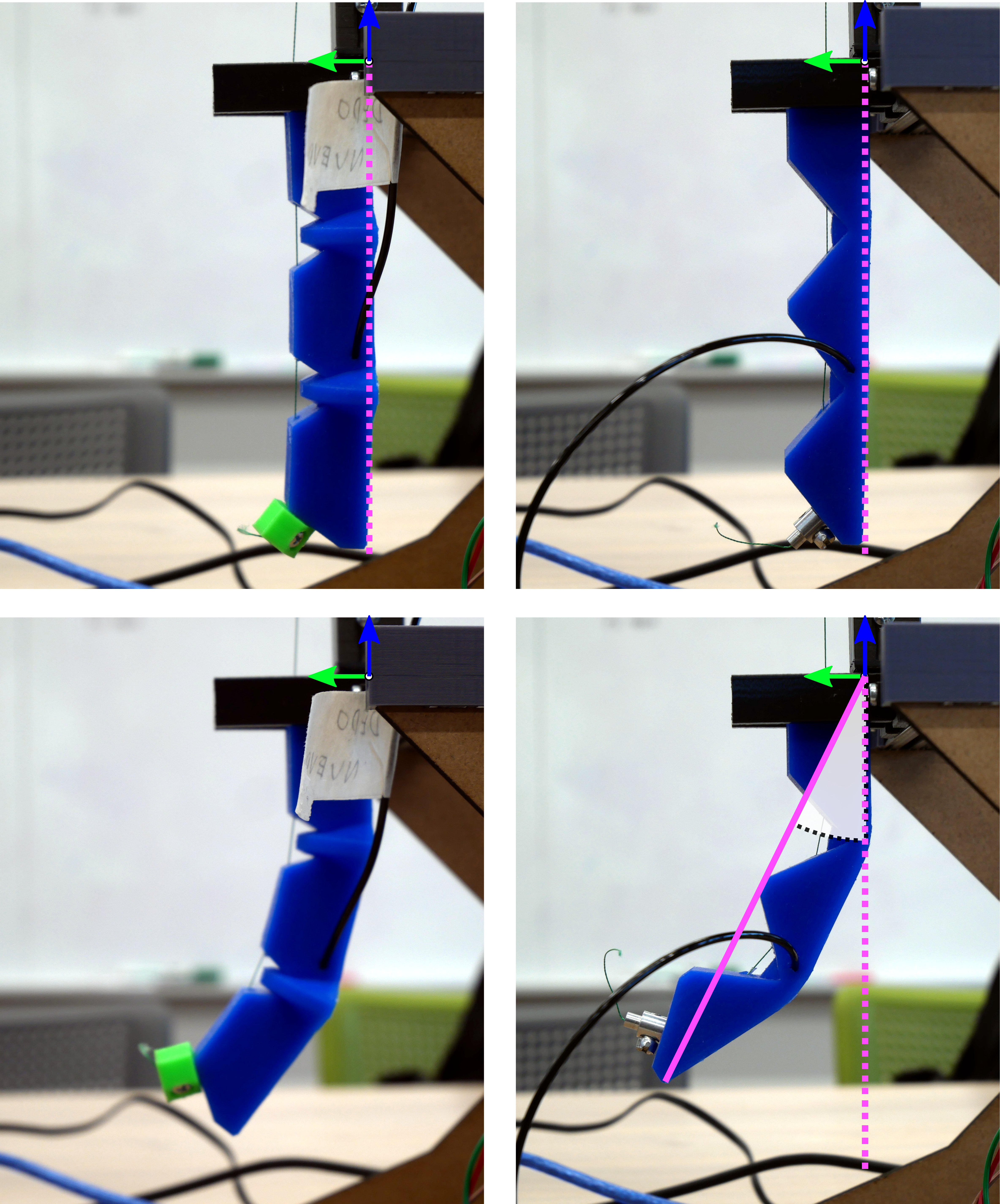}}%
    \put(0.85511437,0.37271386){\color[rgb]{0,0,0}\rotatebox{90}{\makebox(0,0)[lt]{\lineheight{1.25}\smash{\begin{tabular}[t]{l}$\scriptstyle 26.1\si{\degree}$\end{tabular}}}}}%
    \put(0,0){\includegraphics[width=\unitlength,page=2]{FingerDesignComparison.pdf}}%
    \put(0.3614974,0.36995083){\color[rgb]{0,0,0}\rotatebox{90}{\makebox(0,0)[lt]{\lineheight{1.25}\smash{\begin{tabular}[t]{l}$\scriptstyle  16.9\si{\degree}$ \end{tabular}}}}}%
    \put(0,0){\includegraphics[width=\unitlength,page=3]{FingerDesignComparison.pdf}}%
  \end{picture}%
\endgroup%

\caption{Comparison of angular displacement between the large finger (left) and the slim finger (right).}
\label{fig:ComparisonAngularDisplacement}
\end{figure}

To evaluate the performance of the designs in terms of angular displacement, we have manually labelled pictures of the angular displacements with respect to the rest position of both the slim and large fingers, as seen in Fig.~\ref{fig:ComparisonAngularDisplacement}. The results have been summarized in Table~\ref{tab:AngularDisplacement}. The angular displacement is measured with respect to the vertical line (the $\hat{z}$-axis) and the reference point on the finger is at the back of its tip, as shown in Fig.~\ref{fig:ComparisonAngularDisplacement}. There is a very good agreement between the predicted difference in angular displacement of both designs. However, the absolute values differ slightly. Overall these results confirm that the simulation is able to predict the difference in performance between the designs.

\begin{table}
\centering
\begin{tabular}{||c|c|c||} 
 \hline
 % Device & Vol.\ Sensitivity (\si{\micro\liter}) & Std (\si{\micro\liter}) & Std (\%) \\ 
 Angular Displacement (deg.) & Reality & SOFA \\ [0.5ex] 
 \hline
 \hline
 Finger Slim  & 26.1 & 28.13 \\
 \hline
 Finger large &  16.9 &  19.1 \\ 
\hline 
Diff & 9.2 &  9.03 \\
 \hline
 \hline 
\end{tabular}
\caption{Comparison Angular Displacement}
\label{tab:AngularDisplacement}
\vspace{-5mm}
\end{table}

\subsection{Considerations on Sim-to-Real Transfer}
In this sub-section, we dig deeper into the question of how well the results obtained with the help of the simulation transfer to reality. The main points addressed are the mesh density employed, the mechanical parameters chosen (Poisson's Ratio and Young's Modulus) as well as the influence of tolerances due to fabrication. 

\subsubsection{Mesh Density and Deformation Volumes}
\label{subsubsec:MeshDensity}

\begin{figure}
    \centering
    \includegraphics[width=0.7\linewidth]{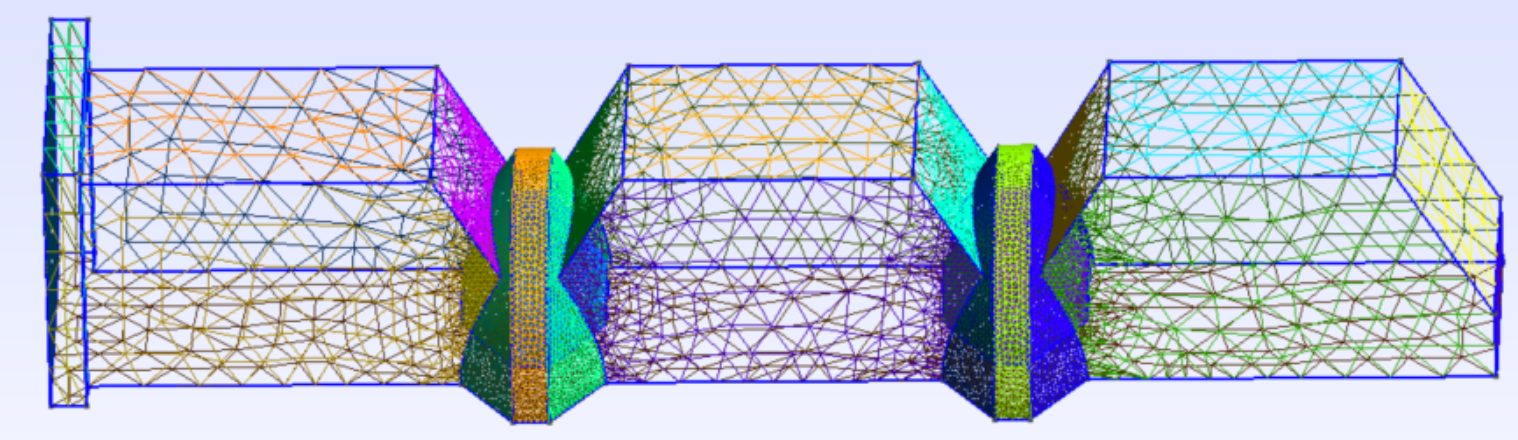}
    \caption{Local mesh size }
    \label{fig:LocalMeshSize}
\end{figure}

\begin{figure}
\centering
 \resizebox{0.9\linewidth}{!}{%
\includegraphics{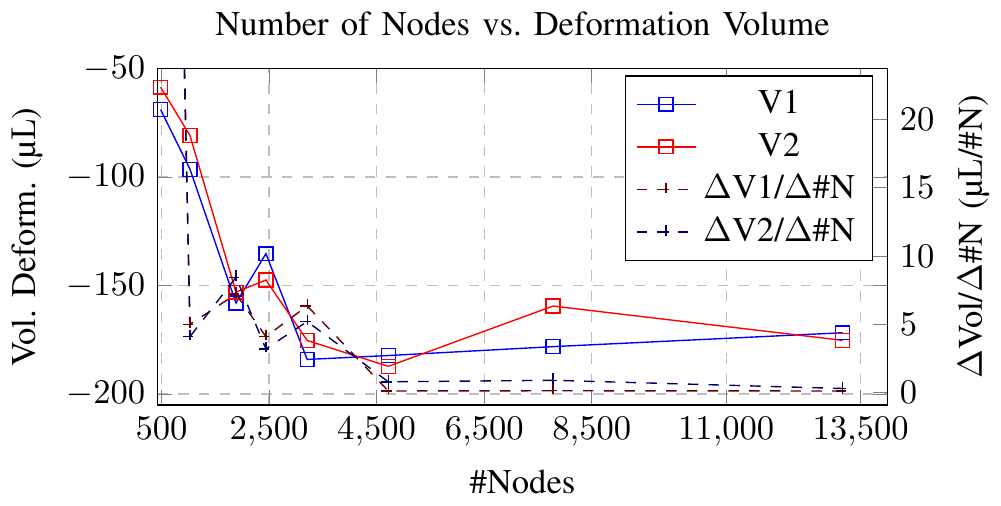}}
\caption{The effect of the number of nodes chosen for representing the finger in SOFA.}
\label{fig:NodesVsSensitivity}
\end{figure}

We used Gmsh's ability to locally define mesh sizes to selectively increase the mesh density around the cavities, as shown in Fig.~\ref{fig:LocalMeshSize}. We observed that increasing the mesh density can affect the estimate of volume change significantly. We therefore repeated the same simulation increasing each time the mesh density to observe this effect. The results are shown in Fig.~\ref{fig:NodesVsSensitivity}.  A higher mesh density captures more detail of the deformation around the cavity and the overall trend is for the sensitivity to increase. However, as the number of nodes employed increases, the estimate converges, as the variation in sensitivity per increase in number of nodes, i.\,e.\ $\Delta V/\Delta \#N$ converges to 0. 

As discussed later, in Sec.~\ref{subsubsec:ParetoFrontVariability}, the Pareto Front for our optimization problem is not very sensitive to the mesh size. Therefore, lower resolutions can be chosen here. Furthermore, this analysis points to the issue of the sim-to-real gap as deformation volumes are not predicted exactly, even for high mesh densities. Possible explanations include that we are ignoring local self-contacts at the folds and tolerances in the fabrication process (see Sec.~\ref{subsubsec:FabricationTolerances}). It should also be considered that the estimated deformation volumes are less than 10\% of the total cavity volume. Overall, empiric calibration factors or mappings are needed to predict the real world values from the simulation if exact predictions are needed.

\subsubsection{Pareto Front Variability}
\label{subsubsec:ParetoFrontVariability}

\begin{figure}
    \centering
    \small
    \includegraphics[width=250pt]{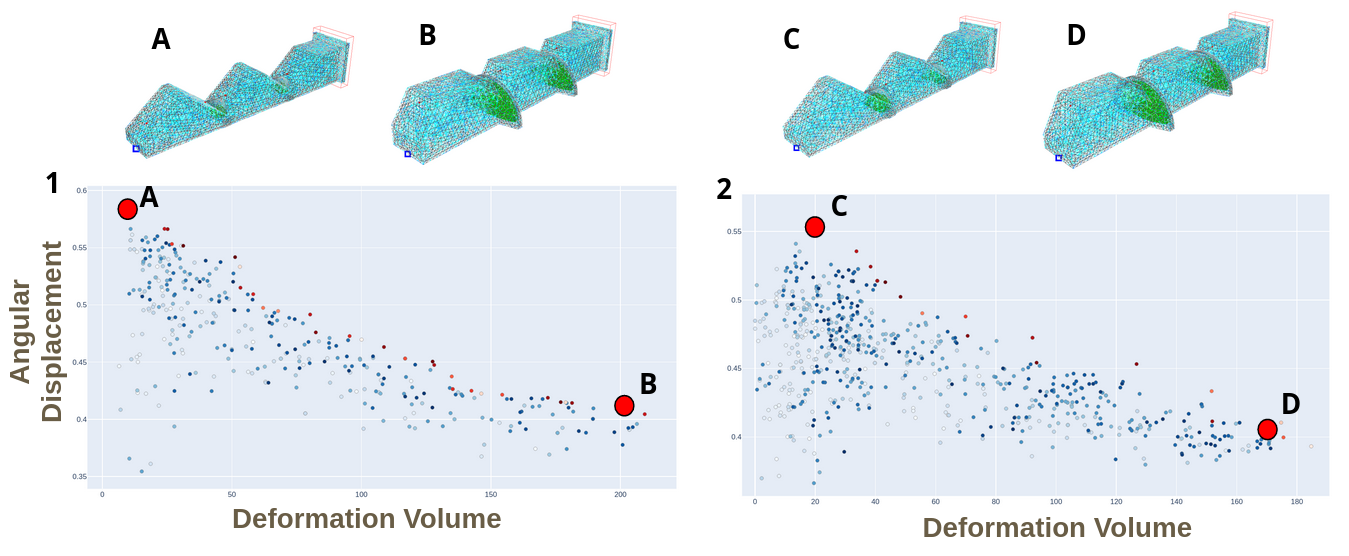}
     \caption{Pareto Fronts as well as optimal geometries obtained regarding both deflection and Deformation Volume for refined meshes and Poissons's Ratios values of 1) $0.45$ with 2500 nodes and 2) $0.495$ with 500 nodes.
     Simulation times range from 2 seconds per design with meshes of 500 nodes to 15 seconds for meshes of 2500 nodes on our computational setup.}
\label{fig:ParetoFrontVariability}
\end{figure}

At first, when exploring the optimal design of the soft finger in Sec.~\ref{subsec:OptimizationResults}, we made two strong assumptions without prior calibration with a physical prototype as stated there, which were a Young's Modulus of \SI{3}{\mega\pascal}, a Poisson's Ratio of 0.3 and a mesh density of 500 nodes (see also Fig.~\ref{fig:ParetoFrontVariability}). To investigate the variability of the Pareto Front, we ran two optimization procedures, once with a Poisson's Ratio of 0.45, having a mesh density of 2500 and once with a Poisson's Ratio of 0.495 and a mesh density of 500 nodes. Here, we show that the Pareto Fronts obtained, shown in Fig.~\ref{fig:ParetoFrontVariability}, are not very dependent on these two parameters. Points on the extremes of the compromises are basically identical designs on either Pareto Front and correspond to the finger slim and finger large. This demonstrates that using a lower resolution mesh and non-calibrated mechanical parameters to explore our Pareto Front is feasible. Nevertheless, this should not be taken for granted and would need to be verified for other optimization tasks. 

Note that we do not consider here the Young's Modulus because as it has little impact on the result. This is due to the performance metrics, the deformation volume of the cavities and angular displacement, which are geometric in nature. An assessment of the finger, for example taking into account contacts in the optimization problem, would have required a prior calibration of this parameter. The same is true for an actuation based on forces on the cable rather than displacement, which would also have required accurate mechanical parameters calibration.

\subsubsection{Fabrication Tolerances}
\label{subsubsec:FabricationTolerances}
\begin{figure}
\centering
 \resizebox{0.9\linewidth}{!}{%
\includegraphics{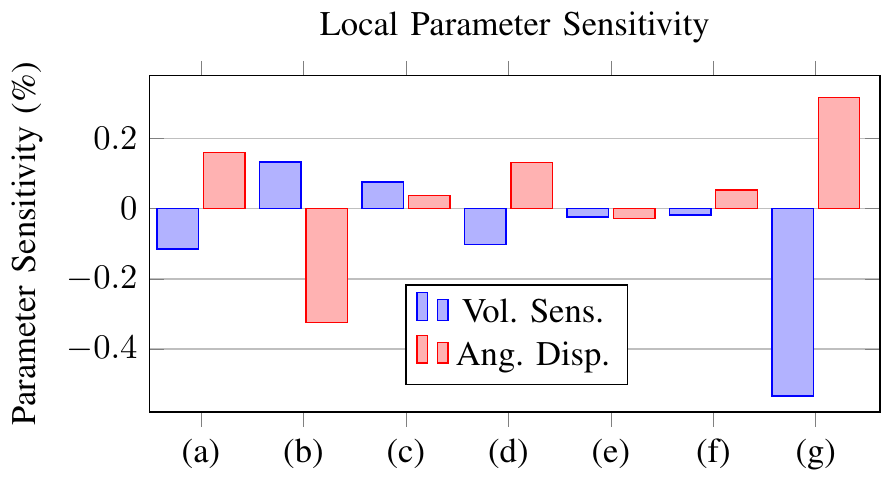}}
\caption{Analysis of local parameter changes around the large finger design using the One-At-a-Time strategy. The labels correspond to the parameters described in Fig.~\ref{fig:FingerDrawing}. The values for each label with respect to each objective correspond to the maximum objective differentials encountered by separately varying each parameter with respect to the initial parameters of the large Finger. These values are then normalized by the maximum objective differential encountered by varying all design variables.}
\label{fig:SensitivityAnalysis}
\end{figure}

We also are interested in studying the effect that fabrication tolerances can have on the performance metrics. Since our process of fabrication is based on 3D-printed molds, we expect tolerances in the fabricated devices related to this process, such as larger than intended borders. These in turn can result in thinner than expected features in the silicone casting process. We have looked at the parameter \emph{Wall Thicknes}, (g) in Fig.~\ref{fig:FingerDrawing}, of the large finger and we measured that it exhibited a variation of up to $\SI{0.4}{\milli\meter}$, going down from the intended $\SI{3}{\milli\meter}$ to $\SI{2.6}{\milli\meter}$. As shown in Table~\ref{tab:FabricationSensitivity}, when simulating the device with the adjusted \emph{Wall Thickness} parameter, using 3209 nodes (see Fig.~\ref{fig:NodesVsSensitivity}), there is a significant impact on the deformation volume, an increase of about 24\%. This is an important reduction of the sim-to-real gap.  We have provided a \emph{sensitivity analysis} for the deformation volume with respect to the free design parameters shown in Fig.~\ref{fig:SensitivityAnalysis}.
%\TN{We use a One-At-a-Time strategy as described in Eq.~\ref{eq:sensitivityOAaT} around the finger large design (see Fig.~\ref{fig:FingerDrawing})
We used a One-At-a-Time strategy. Starting from the finger large design (see Fig.~\ref{fig:FingerDrawing}), optimization objectives are evaluated for bounding values of each design variable. This evaluation lacks the consideration of interactions between variables but has the advantage of being quick to evaluate. It is confirmed that the Wall Thickness is indeed a very sensitive parameter for the deformation volume optimization objective. This gives another clue for the calibration procedure: a model calibration could consider fabrication tolerances, for instance by finding the design parameters within a tolerance interval that minimize the sim-to-real difference.

\begin{table}
\centering
\begin{tabular}{||c|c||} 
 \hline
 % Device & Vol.\ Sensitivity (\si{\micro\liter}) & Std (\si{\micro\liter}) & Std (\%) \\ 
 Wall Thickness  & Deform.\ Vol. \\ [0.5ex] 
 \hline
 \hline
 $\SI{3}{\milli\meter}$ & \SI{-175.24}{\micro\liter}\\
 \hline
$\SI{2.6}{\milli\meter}$ &  \SI{-217.00}{\micro\liter} \\ 

 \hline
 \hline 
\end{tabular}
\caption{Analysis Fabrication Tolerances Effect}
\label{tab:FabricationSensitivity}
\vspace{-5mm}
\end{table}

\section{Conclusions and Future Work}
\label{sec:Conclusions}

In this paper, we have presented an open source design optimization toolbox for Soft Robotics. We believe that computational design will become a key technique for designing, sharing and adapting soft robot designs. To illustrate how the toolbox works, we have evaluated its performance on the example of a cable-driven soft finger featuring embedded fluidics sensors. Designs can be generically created and parameterized through scripting with \emph{Python} and \emph{Gmsh} and can then be readily simulated with the open source platform \emph{SOFA}. The finger design task highlights the need for a multi-objective optimization framework, because the optimizing at the same time for actuation capabilities and optimizing for sensing capabilities can be conflicting goals. Thus, a trade-off needs to be found. We provided the details of the toolbox and provide an experimental validation of the proposed optimization procedure. For a given displacement of the cable inside the finger, we chose to optimize the angular deflection achieved as well as the deformation volume measured inside the embedded cavities. It was shown with help of two fabricated devices on the extremes of the Pareto Front, called finger slim and finger large, that the toolbox correctly predicts significant variations of both optimization objectives. We furthermore provide a more in depth analysis of how the simulation parameters chosen affect --or not-- the real-world outcome of the optimization for our task.

In this paper, we have looked at only one example device for the task of design optimization. Thus, for the future, we would like to explore further devices and design objectives. In particular, we would like to explore more complex devices, such as a gripper, whose basic building block could be the finger presented here. Objectives on the kinematics and sensing of the sensorized finger were considered. These objectives were chosen for the ease of measuring their performance on a test bench. This work could be extended to optimizing the design of an entire soft gripper for manipulation by considering additional design parameters, such as the location of the fingers in relation to each other, as well as additional fitness functions evaluating the contact forces and manipulation dexterity of the robot.

In terms of the optimization methods themselves, we would like to work towards more customizable solvers. The toolbox is currently provided with a linking to \emph{Optuna} which sacrifices extensive customization of the optimization algorithms (choosing optimization hyper-parameters, defining interactive design variables constraints, etc.) in favor of ease of use for non-experts in numerical optimization. In the future, we plan to implement more customizable solver libraries.

\ifCLASSOPTIONcaptionsoff
  \newpage
\fi



\bibliographystyle{IEEEtran}
\bibliography{Bibliography.bib}


% \end{thebibliography}



%\vfill


\end{document}